\documentclass[cameraready]{Interspeech}

\usepackage{cleveref}
\usepackage{multirow}
\usepackage{adjustbox}
\usepackage{makecell}
\usepackage{cite}
\usepackage{siunitx}

\usepackage{amsmath}
\newcommand\numberthis{\addtocounter{equation}{1}\tag{\theequation}}

\usepackage{amssymb}
\usepackage{pifont}
\newcommand{\cmark}{\ding{51}}%
\newcommand{\xmark}{\ding{55}}%

\makeatletter
\renewcommand{\paragraph}[1]{\par\addvspace{0.5em \@plus1ex \@minus.2ex}\noindent\textbf{#1}\quad\ignorespaces}
\makeatother

\title{Text-Utilization for Encoder-dominated Speech Recognition Models}

\author[affiliation={1,2}, orcid=0000-0002-6655-671X]{Albert}{Zeyer}
\author[affiliation={1}]{Tim}{Posielek}
\author[affiliation={1,2}, orcid=0000-0003-2839-9247]{Ralf}{Schlüter}
\author[affiliation={1,2}]{Hermann}{Ney}

\address{
$^1$Machine Learning and Human Language Technology Group, RWTH Aachen University, Germany \\
$^2$AppTek, Germany
}

\email{\{zeyer,schlueter,ney\}@cs.rwth-aachen.de, tim@basetec.de}

\keywords{Speech Recognition, Text Utilization, Encoder-dominated Models, Dynamic Downsampling}

\begin{document}

\maketitle

\global\csname @topnum\endcsname 0
\global\csname @botnum\endcsname 0

\begin{abstract}
This paper investigates efficient methods for utilizing text-only data to improve speech recognition,
focusing on encoder-dominated models that facilitate faster recognition.
We provide a comprehensive comparison of techniques to integrate text-only data,
including modality matching and dynamic downsampling
to reach text-level representations within the encoder.
Our experiments on the LibriSpeech corpus show that a larger encoder with a smaller decoder
can equal or surpass the performance of architectures with larger decoders.
We demonstrate that simple configurations,
such as random duration models,
are often more effective than complex alternatives,
significantly simplifying the training pipeline.
All code and recipes are made publicly available.
\end{abstract}

\section{Introduction}

\begin{figure}
\centering
\includegraphics[width=0.4\linewidth]{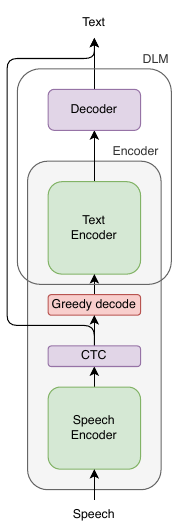}
\caption{Speech recognition using denoising LM \cite{gu2024lm,koch2025dlm}.}
\label{fig:dlm}
\end{figure}

\begin{figure}
\centering
\includegraphics[width=\linewidth]{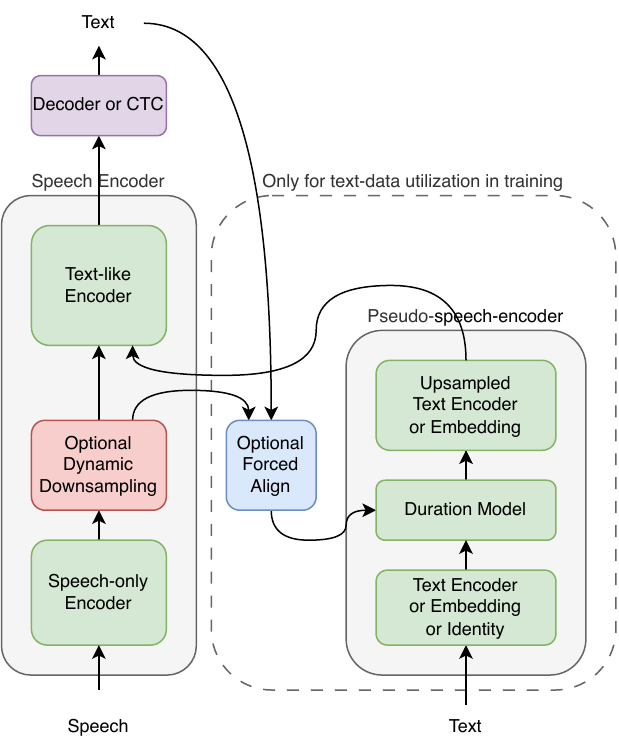}
\caption{Model architecture including the training part
to utilize the text-only data, which is not used during recognition.}
\label{fig:model}
\end{figure}

It remains an open question how to best utilize the large amount of text-only data for improving ASR,
in addition to the paired audio-text data.
A standard approach is to train a separate language model (LM) on the text-only data,
and then use shallow fusion to utilize the LM during decoding.
Using text-to-speech (TTS) to convert the text-only data into synthetic speech and then train on that together with the paired audio-text data
is another approach, which has been shown to be very effective
\cite{rossenbach:ICASSP2020,gu2024lm,koch2025dlm}.

We summarize these baselines in \Cref{tab:baselines}.
These results lead to the main motivation of this paper,
that a smaller decoder with a larger encoder is just as good
or even better than a large decoder with a smaller encoder.
Using TTS to utilize the text-only data directly during the ASR training
is a very strong baseline,
outperforming the separate LM training and shallow fusion approach.
Using both TTS + LM together still gains further.
Replacing the LM by a denoising language model (DLM)
\cite{gu2024lm,koch2025dlm}
still results in small further improvements.
However, the interesting observation is that the CTC + DLM
has a smaller decoder and a larger encoder than the CTC + LM,
which is more efficient for recognition.
Similarly, a simple CTC + AED model with a smaller decoder and a larger encoder
results in almost the same performance as the CTC + DLM,
even though the training and recognition is much simpler,
just utilizing TTS for the text-only data.

We note that the DLM is an AED model operating on text,
on top of the CTC output.
Thus the joint CTC + DLM model is like an extended encoder with downsampling in between to get to text-level representation,
then another text-level encoder on top of that, which utilized the text data,
and a decoder on top of that,
as shown in \Cref{fig:dlm}.
It was shown in \cite{koch2025dlm} that the decoder can be very small in relation to the encoder,
which is what we also argue here.
Note that the TTS data utilized the text data on first part of the encoder
and also the DLM.

These results show that a larger encoder with a smaller decoder is just as good or even better than a smaller encoder with a larger decoder.
With a larger encoder, it is crucial that the text data is utilized in the encoder as well,
not only in the decoder.
Here we explore different methods that avoid the TTS approach,
which is computationally expensive and also requires a separate TTS model to be trained.
The CTC + AED training on TTS data, while very simple and leading to very good results,
is very expensive to train compared to the CTC on the paired audio-text data only
and separate LM training on the text-only data,
since the TTS representation of the text data is much larger than the text representation of the text data.
Thus, our goals:
\begin{itemize}
\item We want a simple training and recognition approach that can utilize the text-only data effectively.
Thus, we want to avoid relying on a TTS model.
The DLM approach also requires the TTS.

\item We want to improve the efficiency of training while utilizing the text-only data.
While the CTC + AED training on TTS data is simple (despite the TTS), it is very expensive to train.

\item We want to have models that are efficient in recognition,
which means encoder-dominated models with a small decoder.

\end{itemize}
Our contributions are as follows:
\begin{itemize}
\item We point out that an encoder-dominated model (larger encoder with a smaller decoder)
is just as good or even better than a decoder-dominated model (smaller encoder with a larger decoder)
when utilizing the text-only data
(\Cref{tab:baselines}).

\item For encoder-dominated models, it is crucial to utilize the text-only data in the encoder as well
(\Cref{tab:encoder-decoder-training}).

\item We study different dynamic downsampling variants in the middle of the encoder to get to text-level representation,
to allow for more efficient training,
and easier utilization of the text-only data in the upper text-level part of the encoder.

\item We compare various approaches to utilize the text-only data in the encoder
and various variants of a pseudo-speech-encoder.
This provides a unified comparison of existing text-utilization strategies
such as modality matching
(similar to MAESTRO \cite{chen2022maestro}, CJST \cite{zhou24cjst})
and direct encoder training
(similar to JOIST \cite{sainath2023joist}, SpeechUT \cite{zhang2022speechut}, SpeechT5 \cite{ao-etal-2022-speecht5})
under a single, controlled framework.
\end{itemize}

\section{Related Work}

A lot of work is on utilizing a pre-trained language model (LM) in the decoder,
or even a large LM (LLM).
The (L)LM is extended to integrate the encoder,
usually by prepending the encoder output adapted into the LM dimension,
known as prefix-LM \cite{wang20222prefix-lm},
and then finetuned on paired audio-text data.


Text-utilization in encoder
has also been explored in various ways,
such as
modality matching
\cite{chung-etal-2021-splat,chen2022maestro,wang2023understandingsharedrepr,%
zhou24cjst,gaur24astra},
direct utilization of the text data as input to the encoder
\cite{bapna2021slamunifiedencoderspeech,bapna2022mslammassivelymultilingualjoint,tang-etal-2022-unified,thomas2022textogram,ao-etal-2022-speecht5,zhang2022speechut,sainath2023joist,Sunder2025nonautoregressive}
or TTS
\cite{rossenbach:ICASSP2020,dalterio2023tts,gu2024lm,koch2025dlm}.


Downsampling to text-level representation in the middle of the encoder
via CTC compression has also been explored
\cite{gao22b_interspeech,an2024paraformerv2,zou24_interspeech}.


\section{Model}
\label{sec:model}

The model architecture is shown in \Cref{fig:model}.

\paragraph{Speech-only encoder.}
We perform log-mel feature extraction on the input audio,
then a convolutional front-end to downsample the features by a factor of 6 in time,
and then our speech encoder which is a stack of Conformer blocks.

\paragraph{Downsampling to text-level.}
After the speech encoder, we optionally perform dynamic downsampling
via a variant of CTC compression \cite{gaido-etal-2021-ctc,zhou24cjst},
utilizing an intermediate CTC output to determine which frames to merge together,
to get a text-level-like representation.
%
\begin{align*}
    \hat{y}_t &= \operatorname{argmax}_y p_{\textrm{iCTC}}(y \mid h_t), \\
    m_t &= (\hat{y}_t = \hat{y}_{t-1}) && \text{(merge mask)} \\
    &\quad \wedge (p_{\textrm{iCTC}}(\hat{y}_t \mid h_t) \ge \tau) \\
    &\quad \wedge (p_{\textrm{iCTC}}(\hat{y}_{t-1} \mid h_{t-1}) \ge \tau), \\
    i_t &= \begin{cases}
        1 & \text{if } t = 1 \\
        i_{t-1} & \text{if } m_{t}, t \ge 2 \\
        i_{t-1} + 1 & \text{if } \neg m_{t}, t \ge 2
    \end{cases} \numberthis \label{eq:ctc-compression-merged-index}
\end{align*}
where $\tau$ is a threshold hyperparameter.
Then we can compress the encoder output by pooling over the frames with the same index $i_t$:
\begin{align*}
    \operatorname{compress}(h_1^T)_i &= \operatorname{mean}(\{ h_t \mid i_t = i \}), \quad i \in 1, \ldots, i_T .
\end{align*}

\paragraph{Text-like encoder.}
After the downsampling,
we have a text-like encoder which is a stack of
Transformer++ or Conformer blocks.

\paragraph{Output.}
CTC on output.
Small decoder with cross-attention.

\subsection{Pseudo-speech-encoder}

The pseudo-speech-encoder is a module that converts text input into pseudo-speech representation,
which is then fed into the text-like encoder,
to allow the text data to be utilized in the encoder as well.
This part is only used for training with unpaired text-only data,
and is not used during recognition.

It starts with an embedding layer, or with interleaving blanks between the text tokens and then embedding.
An optional text encoder follows.
After that,
we have either a trainable duration model,
or a non-trainable purely random duration model.
Finally, we have a linear layer to convert the dimension
into the same as the text-like encoder dimension.

\subsection{Recognition}

We apply time-synchronous beam search decoding
using the CTC output from the speech encoder
and the decoder output.

\section{Training}

\subsection{Paired audio-text data}
We use the typical supervised training with cross-entropy loss on the decoder output,
and CTC loss on the encoder output.

Trainable duration models in the pseudo-speech-encoder
are trained with the paired audio-text data,
by using the CTC forced alignment
between the speech encoder output and the text encoder output.

\subsection{Unpaired text-only data}

\paragraph{Pseudo-speech-encoder.}
The data is fed into the pseudo-speech-encoder to get the pseudo-speech representation,
which is then fed into the text-like encoder.
All the losses on top of the text-like encoder and decoder
are the same as in the paired audio-text data training,
i.e., the same cross-entropy loss on the decoder output,
and the same CTC loss on the text-like encoder output.

\paragraph{TTS.}
As an extreme variant,
we also use a TTS model to convert the text input into synthetic speech,
and then feed the synthetic speech into the speech encoder.
In this case, we do not have the pseudo-speech-encoder at all.

\subsection{Pseudo-speech-encoder training variants.}
We have two variants:

\paragraph{Modality matching.}
The pseudo-speech-encoder is trained in a modality matching way with the speech encoder:
We use a CTC forced alignment to apply the durations on the text input
exactly such that the pseudo-speech representation length matches the speech encoder output length.
Then we use simple L2 loss to match the pseudo-speech representation with the speech encoder output.
In the unpaired text-only data training,
we stop the gradient from the text-like encoder and decoder to the pseudo-speech-encoder,
so that the pseudo-speech-encoder is only trained with the paired audio-text data using the modality matching loss.

\paragraph{Direct.}
We don't train the pseudo-speech-encoder with any modality matching loss.
In the paired audio-text data training,
the pseudo-speech-encoder is not trained at all
(except for the duration model if it is trainable).
In the unpaired text-only data training,
we allow the gradient to backpropagate to the pseudo-speech-encoder,
so that it is trained jointly with the text-like encoder and decoder.

\subsection{Batching and data ratios}

We have three cases:

\paragraph{Only paired audio-text data.}
This is the typical supervised training.
We use no unpaired text-only data at all in this case.
We use no gradient accumulation.

\paragraph{Unpaired text-only data utilized via TTS.}
The unpaired text-only data is converted into synthetic speech by the TTS model offline and stored as audio files.
Then the training is the same as the paired audio-text data training,
except that we have more data.
We keep an approximate ratio of 1:1 for the paired audio-text data
and the unpaired text-only data (after TTS conversion) during the training,
i.e.~we see the same amount of hours of paired audio-text data
and the synthetic speech from the unpaired text-only data.
We don't use gradient accumulation.

\paragraph{Unpaired text-only data utilized via pseudo-speech-encoder.}
For efficiency reasons,
we always have one batch using only paired audio-text data,
and the other batch using only the unpaired text-only data.
We use gradient accumulation over both batches
such that every update step is based on both types of data.
We vary the total amount of unpaired text-only data used in the training,
and the ratio between the paired audio-text data and the unpaired text-only data.
This effectively varies the batch sizes of the two batches.

\section{Setup and Baselines}

\subsection{Data}

We use the LibriSpeech dataset \cite{panayotov2015librispeech} for our experiments:
We use the standard paired audio-text data for training
consisting of around 960 hours of speech
and the corresponding text transcriptions of around 10M words,
as well as the text-only data from the LibriSpeech LM corpus
consisting of around 800M words,
or around 75k hours of synthetic speech after TTS conversion.

\subsection{Baseline CTC + LM}

The baseline ASR CTC model was trained on 100 epochs of only the paired audio-text data,
i.e. approximately 100k hours of audio data.
We have trained separate language models for 5 epochs on the unpaired text-only data.
Using shallow fusion with the separately trained language model during decoding
provides a strong baseline for utilizing the text-only data.
The result is shown in \Cref{tab:baselines}.

\subsection{Baseline CTC + AED with TTS}

We trained a CTC + AED model
(with shared encoder,
i.e.~like the model described in \Cref{sec:model}
but without downsampling to text-level and without the pseudo-speech-encoder)
on the paired audio-text data with 100 epochs (approximately 100k hours),
as well as the synthetic speech from the unpaired text-only data
with 1.3 epochs (approximately 100k hours) with the same training setup.
The TTS model was trained on the paired audio-text data only,
and follows the public pipeline of \cite{koch2025dlm}.
The result is shown in \Cref{tab:baselines}.

\subsection{Baseline proposed model and method}

All our models use CTC + AED on top.
We always train on 100 epochs of the paired audio-text data,
and on 5 epochs of the unpaired text-only data using the pseudo-speech-encoder.

\section{Experiments}


All our models are CTC + AED models like described in \Cref{sec:model}, \Cref{fig:model},
if not otherwise specified.

\begin{table}
\caption{Baseline results on LibriSpeech,
\label{tab:baselines}
\textbf{comparing different text utilization approaches}
of using the Librispeech text corpus.
Larger encoder with smaller decoder just as good or better.
$^*$Results from \cite{koch2025dlm} utilizing a TTS and denoising language model (DLM).
The TTS in all results here was trained on Librispeech only.
}
\begin{adjustbox}{max width=\linewidth-1ex}
\setlength{\tabcolsep}{3pt}
\begin{tabular}{|l|c|c|c|c|c|c|c|}
\hline
\multirow{3}{*}{Model} & \multicolumn{2}{c|}{\multirow{2}{*}{\shortstack{Num.\\layers}}} &\multirow{3}{*}{\shortstack{Text-\\util.}} & \multicolumn{4}{c|}{WER [\%]} \\
& \multicolumn{2}{c|}{} && \multicolumn{2}{c|}{dev} & \multicolumn{2}{c|}{test} \\
& Enc.& Dec. && clean & other & clean & other \\
\hline
\hline
\multirow{2}{*}{CTC} & \multirow{5}{*}{16} & \multirow{2}{*}{0} & - & 2.27 & 5.08 & 2.42 & 5.32 \\ \cline{4-8}

&& & TTS & 1.76 & 3.98 & 1.90 & 4.15 \\ \cline{1-1} \cline{3-8}

\multirow{2}{*}{CTC + LM} &  & \multirow{2}{*}{32} & LM & 1.83 & 3.94 & 1.99 & 4.26 \\ \cline{4-8}

& &  & \shortstack{TTS\\+LM} & 1.56 & 3.41 & 1.73 & 3.70\makebox[0pt][l]{\hspace{1ex}\footnotesize$*$} \\ \cline{1-1} \cline{3-8}

\multirow{2}{*}{CTC + AED} & & \multirow{2}{*}{6} & - & 1.84 & 4.33 & 2.06 & 4.58 \\ \cline{4-8}

&  &  & \multirow{4}{*}{TTS} & 1.57 & 3.53 & 1.73 & 3.64 \\ \cline{1-3} \cline{5-8}


CTC & \multirow{2}{*}{24} & 0 &  & 1.69 & 3.59 & 1.75 & 3.96 \\ \cline{1-1} \cline{3-3} \cline{5-8}

CTC + AED &  & \multirow{2}{*}{8} &  & 1.49 & 3.31 & 1.60 & 3.56 \\ \cline{1-2} \cline{5-8}

CTC + DLM & 40 &  &  & 1.49 & 3.29 & 1.72 & 3.53\makebox[0pt][l]{\hspace{1ex}\footnotesize$*$} \\ \hline

\end{tabular}
\end{adjustbox}
\end{table}

\subsection{Downsampling \& Text-level Encoder Variants}

\begin{table}
\centering
\caption{\textbf{Effect of self-conditioning} on intermediate CTC outputs \cite{nozaki21_interspeech},
\textbf{effect of downsampling},
and \textbf{effect of text-level encoder type},
with 12 speech-only encoder layers and 4 text-level encoder layers,
and using only paired audio-text data for training (no unpaired text-only data, no TTS).
Using a smaller model with 512 dims.
}
\label{tab:self-conditioning}
\begin{adjustbox}{max width=\linewidth-1ex}
\setlength{\tabcolsep}{2pt}
\begin{tabular}{|c|c|c|c|c|c|c|c|}
\hline
\multirow{3}{*}{\shortstack{Upper\\encoder\\type}} & \multirow{3}{*}{\shortstack{Down-\\sampling}} & \multirow{3}{*}{\shortstack{Self-\\condit-\\ioning}} & \multicolumn{4}{c|}{WER [\%]}
& \multirow{3}{*}{\shortstack{Train\\time\\{[h]}}} \\
&&& \multicolumn{2}{c|}{dev} & \multicolumn{2}{c|}{test}& \\
&&& clean & other & clean & other& \\
\hline\hline


Trafo++ & \xmark & \xmark & 2.24 & 5.54 & 2.35 & 5.71 & 24.7 \\ \cline{3-8}

&& \cmark & 2.30 & 5.52 & 2.37 & 5.93 & 25.6 \\ \cline{2-8}

& \cmark & \xmark & 2.24 & 5.66 & 2.45 & 5.78 & 24.9 \\ \cline{3-8}

&  & \cmark & 2.23 & 5.66 & 2.48 & 5.94 & 24.0 \\ \cline{1-1} \cline{3-8}

Conformer &  & \xmark & 2.24 & 5.55 & 2.41 & 5.75 & 26.5 \\ \cline{3-8}

& & \cmark & 2.18 & 5.61 & 2.41 & 5.63 & 25.7 \\ \hline

\end{tabular}
\end{adjustbox}
\end{table}





Our dynamic downsampling approach is based on the CTC output probabilities,
and the intention is that the upper text-level encoder operates on more text-like representations.
We can encourage this by providing the CTC output probabilities as additional input to the upper text-level encoder.
This is known as \emph{self-conditioning} on intermediate CTC outputs \cite{nozaki21_interspeech}.
\Cref{tab:self-conditioning} showed that is not helpful in our case.
From these results, we also see that the downsampling leads to a slight degradation in WER.
We also see that Conformer layers in the upper text-level encoder are slightly better than Trafo++ layers.

\subsection{Training only the Decoder with Text-only Data}

\begin{table}
\centering
\caption{\textbf{Effect of training the encoder with text-only data}.
The encoder always has 16 layers, 1024 dims, without intermediate downsampling,
and the decoder size is varied.
The encoder text-training uses the simple random duration model, simple pseudo-speech-encoder, and simple text augmentation.
The decoder is always trained with text-only data.
}
\label{tab:encoder-decoder-training}
\begin{adjustbox}{max width=\linewidth}
\setlength{\tabcolsep}{3pt}
\begin{tabular}{|c|c|c|c|c|c|c|}
\hline
\multirow{3}{*}{\makecell{Text\\utilization\\type}} &
\multirow{3}{*}{\makecell{Dec. size\\(num\\layers)}} &
\multirow{3}{*}{\makecell{Text-training\\num enc.~l.\\(out of 16)}} &
\multicolumn{4}{c|}{WER [\%]} \\
&&& \multicolumn{2}{c|}{dev} & \multicolumn{2}{c|}{test} \\
&&& clean & other & clean & other \\
\hline\hline

- & 6 & - & 1.81 & 4.17 & 2.05 & 4.27 \\ \hline

\multirow{4}{*}{\makecell{Dec.\\as\\LM}} & 6 & 0 & 1.89 & 4.40 & 2.10 & 4.56 \\ \cline{2-2}\cline{4-7}

& 12 & & 1.88 & 4.35 & 2.00 & 4.62 \\ \cline{2-2}\cline{4-7}

& 24 & & 1.84 & 4.15 & 2.03 & 4.41 \\ \cline{2-2}\cline{4-7}

& 32 & & 2.14 & 4.17 & 2.10 & 4.62 \\ \hline

\makecell{Pseudo-\\speech-\\enc.} & 6 & 12 & 2.10 & 4.02 & 2.11 & 4.59 \\ \hline

TTS & 6 & 16 & 1.57 & 3.53 & 1.73 & 3.64 \\ \hline

\end{tabular}
\end{adjustbox}
\end{table}

From \Cref{tab:encoder-decoder-training},
we see mostly negative results
when only training the decoder with the text-only data.
Only when also training the encoder with the text-only data, we see a slight improvement.

\subsection{Varying the amount of text-only data ratio}

\begin{table}
\centering
\caption{\textbf{Varying the amount of text-only data ratio}.
The text corpus is the same, but we vary how much we train on it,
how much we repeat it,
by specifying the ratio of text-only data vs.~paired audio-text data.
We always use the pseudo-speech-encoder with the simple random duration model, and the simple text augmentation.
Text ratio 3.75 is the baseline that we also use in the other experiments with pseudo-speech-encoder.
Encoder has 16 layers and 1024 dims, and the decoder has 6 layers, and we use AED+CTC.
}
\label{tab:txt-ratio}
\begin{adjustbox}{max width=\linewidth}
\setlength{\tabcolsep}{3pt}
\begin{tabular}{|S[table-format=1.2]|S[table-format=1.2]|S[table-format=1.2]|S[table-format=1.2]|S[table-format=1.2]|S[table-format=3.1]|}
\hline
{\multirow{3}{*}{\makecell{Text\\ratio}}}
& \multicolumn{4}{c|}{WER {[\%]}}
& {\multirow{3}{*}{\shortstack{Train\\time\\{[h]}}}} \\
& \multicolumn{2}{c|}{dev} & \multicolumn{2}{c|}{test} & \\
& {clean} & {other} & {clean} & {other} & \\
\hline\hline

0.0 & 1.81 & 4.17 & 2.05 & 4.27 & \phantom{0}78.2 \\ \hline


1.0 & 1.85 & 4.20 & 2.08 & 4.54 & 120.0 \\ \hline

1.5 & 1.88 & 4.14 & 2.10 & 4.59 & 135.8 \\ \hline

3.75 & 2.10 & 4.02 & 2.11 & 4.59 & 238.0 \\ \hline

\end{tabular}
\end{adjustbox}
\end{table}

We vary the text-only data ratio in \Cref{tab:txt-ratio}.
We see that the best WER is achieved with the highest text ratio,
which is also the one we use in the other experiments with pseudo-speech-encoder.


\subsection{Simple vs.~complex Pseudo-Speech-Encoder}

\begin{table}
\caption{Comparison of \textbf{pseudo-speech-encoder components}:
input before applying duration model (simply interleaving blank labels, or small Transformer + transposed convolution),
\textbf{duration model} (random or trained),
upsampled text encoder (small Transformer or just embedding or nothing),
and whether we do \textbf{modality matching} (MSE loss) or not.
The speech encoder is with intermediate downsampling,
and the model is smaller with 512 dims.
\label{tab:interleaver_duration_outencoder}
}
\begin{adjustbox}{max width=\linewidth-1ex}
\setlength{\tabcolsep}{3pt}
\begin{tabular}{|c|c|c|c|c|c|c|c|}
\hline
\multicolumn{4}{|c|}{Pseudo-speech-encoder} & \multicolumn{4}{c|}{WER {[\%]}} \\
\multirow{2}{*}{\shortstack{Input}} & \multirow{2}{*}{\shortstack{Duration\\model}} & \multirow{2}{*}{\shortstack{Upsmpl.\\encoder}} & \multirow{2}{*}{\shortstack{Modal.\\match.}} & \multicolumn{2}{c|}{dev} & \multicolumn{2}{c|}{test} \\
&&&& clean & other & clean & other \\
\hline
\hline






\multirow{4}{*}{\shortstack{Inter-\\leave\\blanks}}
& \multirow{1}{*}{random}
& Embed
& \multirow{1}{*}{\xmark}
& 2.18 & 5.35 & 2.40 & 5.68 \\ \cline{2-2} \cline{4-8}


& \multirow{1}{*}{trained}
&
& \multirow{1}{*}{\cmark}
& 2.31 & 5.59 & 2.35 & 5.69 \\ \cline{2-3} \cline{5-8}

& \multirow{1}{*}{random}
& Trafo
&
& 2.31 & 5.51 & 2.38 & 5.64 \\ \cline{2-2} \cline{5-8}

& trained &  & & 2.30 & 5.78 & 2.47 & 5.72 \\ \hline


\multirow{4}{*}{\shortstack{Trafo +\\Transp.\\Conv}}
& \multirow{1}{*}{random}
& -
& \xmark & 2.29 & 5.49 & 2.42 & 5.55 \\ \cline{2-2} \cline{4-8}

& trained & & \cmark & 2.25 & 5.69 & 2.47 & 5.65 \\ \cline{2-3} \cline{5-8}

& random & Trafo &  & 2.28 & 5.65 & 2.45 & 5.69 \\ \cline{2-2} \cline{5-8}

& trained & &  & 2.31 & 5.51 & 2.43 & 5.83 \\ \hline

\end{tabular}
\end{adjustbox}
\end{table}






\begin{table}
\centering
\caption{
Compare different pseudo-speech-encoder
\textbf{output masking and input text augmentation} options.
Using random duration
and simple pseudo-speech-encoder (just embedding layer).
}
\label{tab:augmentation_vs_masking}
\begin{adjustbox}{max width=\linewidth-1ex}
\setlength{\tabcolsep}{3pt}
\begin{tabular}{|c|c|c|c|c|c|}
\hline
\multirow{3}{*}{Masking} & \multirow{3}{*}{\shortstack{Text\\Augmentation}} & \multicolumn{4}{c|}{WER [\%]} \\
&& \multicolumn{2}{c|}{dev} & \multicolumn{2}{c|}{test} \\
&& clean & other & clean & other \\
\hline
\hline

0.2 & 0.0 & 2.23 & 5.30 & 2.33 & 5.60 \\
\hline

0.0 & 0.2 & 2.21 & 5.36 & 2.36 & 5.38 \\
\hline

0.2 & 0.2 & 2.22 & 5.31 & 2.38 & 5.48 \\
\hline

\end{tabular}
\end{adjustbox}
\end{table}

\begin{table}
\centering
\caption{Variation of the \textbf{maximal duration of blanks and labels} for random duration prediction.
\label{tab:blank_labal_max}
}
\begin{adjustbox}{max width=\linewidth}
\setlength{\tabcolsep}{3pt}
\begin{tabular}{|c|c|c|c|c|c|}
\hline
\multirow{3}{*}{\makecell{Max.\\blank\\dur.}} &
\multirow{3}{*}{\makecell{Max.\\label\\dur.}} &
\multicolumn{4}{c|}{WER [\%]} \\
&& \multicolumn{2}{c|}{dev} & \multicolumn{2}{c|}{test} \\
&& clean & other & clean & other \\
\hline
\hline
0 & 1 & 2.26 & 5.55 & 2.36 & 5.50 \\
\hline

1 & 1 & 2.20 & 5.28 & 2.39 & 5.61 \\
\hline

2 & 1 & 2.17 & 5.44 & 2.35 & 5.47 \\
\hline

3 & 1 & 2.22 & 5.31 & 2.38 & 5.48 \\
\hline

3 & 2 & 2.23 & 5.44 & 2.37 & 5.68 \\
\hline

\end{tabular}
\end{adjustbox}
\end{table}

The pseudo-speech-encoder has several components that can be varied,
such as the input (interleaving blanks or using a small text encoder + transposed convolution),
the duration model (random or trained),
the upsampled text encoder (small Transformer or just embedding or nothing),
and whether we do modality matching (MSE loss) or not.
This reduces to the various approaches from the literature
such as MAESTRO \cite{chen2022maestro} (interleaving blanks, learned durations, simple Transformer, explicit modality matching).
We compare each of these components in \Cref{tab:interleaver_duration_outencoder}.
We see that the simpler approach of just interleaving blanks and a simple embedding layer on top of random durations
without modality matching
actually performs better
than the more complex approach
with a small text encoder, trained duration model, and modality matching.
In most cases, a random duration model performs better than a trained duration model.
Explicit modality matching does not seem necessary or helpful in our experiments.
%
We compare different masking and augmentation options in \Cref{tab:augmentation_vs_masking}
and observe only minor differences.
We also compare different maximal durations for blanks and labels in the random duration model in \Cref{tab:blank_labal_max}
and see that the differences are minor,
which allows us to use shorter durations and thus faster training.

\subsection{Amount of encoder that utilizes text-only data}

\begin{table}
\centering
\caption{
Comparing \textbf{different splits of speech-only encoder layers and text-level encoder layers},
with a total of 16 layers for the speech encoder.
There is no downsampling.
We utilize the unpaired text-only data for training
via simple pseudo-speech-encoder
with random duration model.}
\label{tab:SpeechLayerSplit}
\begin{adjustbox}{max width=\linewidth}
\setlength{\tabcolsep}{3pt}
\begin{tabular}{|c|c|c|c|c|c|}
\hline
Speech Layers & Text Layers & \multicolumn{4}{c|}{WER [\%]} \\
&& \multicolumn{2}{c|}{dev} & \multicolumn{2}{c|}{test} \\
&& clean & other & clean & other \\
\hline
\hline
0 & 16 & 2.20 & 5.32 & 2.43 & 5.67 \\
\hline

3 & 13 & 2.16 & 5.18 & 2.29 & 5.41 \\
\hline

4 & 12 & 2.18 & 5.10 & 2.23 & 5.41 \\
\hline

6 & 10 & 2.10 & 5.27 & 2.30 & 5.31 \\
\hline 

8 & 8 & 2.16 & 5.20 & 2.32 & 5.41 \\
\hline

\end{tabular}
\end{adjustbox}
\end{table}

In \Cref{tab:SpeechLayerSplit} we compare different splits of speech-only encoder layers and text-level encoder layers,
and see that having 4 speech-only encoder layers and 12 text-level encoder layers works best,
but the differences are not very large.

\subsection{CTC-only Results}

The text-utilization on the encoder side
shows its advantage on a pure CTC model,
without the decoder,
to allow for very efficient non-autoregressive decoding.
Our results in \Cref{tab:ctc-results-txt-util}
show some improvement with the pseudo-speech-encoder
over not using text-utilization.
But we also see that there is still a significant gap to the TTS-based text-utilization.

\begin{table}
\centering
\caption{\textbf{Effect of text-utilization on pure CTC model}.
The encoder has 16 layers, 1024 dims in total.
The encoder text-training with pseudo-speech-encoder uses the simple variant with random duration model,
feeding it into the encoder after layer 4.
The TTS text-utilization training time does not include the time for training the TTS model
nor for the synthesis of the TTS data.
}
\label{tab:ctc-results-txt-util}
\begin{adjustbox}{max width=\linewidth}
\setlength{\tabcolsep}{5pt}
\begin{tabular}{|c|c|c|c|c|c|c|}
\hline
\multicolumn{2}{|c|}{\multirow{2}{*}{\makecell{Text\\utilization}}} &
\multicolumn{4}{c|}{WER [\%]}
& \multirow{3}{*}{\shortstack{Train\\time\\{[h]}}} \\
\multicolumn{2}{|c|}{} & \multicolumn{2}{c|}{dev} & \multicolumn{2}{c|}{test} & \\
Type & \makecell{Text\\ratio} & clean & other & clean & other & \\
\hline\hline


- & 0.0\phantom{0} & 2.21 & 4.87 & 2.43 & 5.14 & \phantom{0}78.2 \\ \hline

\makecell{Pseudo-\\speech-enc.} & 3.75 & 2.01 & 4.50 & 2.22 & 4.95 & 238.0 \\ \hline

TTS & 1.33 & 1.76 & 3.98 & 1.90 & 4.15 & 176.2 \\ \hline

\end{tabular}
\end{adjustbox}
\end{table}

\section{Conclusions \& Future Work}

We have shown that encoder-dominated models are highly effective for speech recognition
when text-only data is utilized within the encoder.
Our results indicate that a larger encoder paired with a small decoder
provides performance that is equal to or better than models with larger decoders (\Cref{tab:baselines}).
In encoder-dominated models,
text-utilization on the encoder-side training is essential.
The proposed pseudo-speech-encoder enables the integration of large-scale text data
without the high computational cost of TTS.
Furthermore, we found that simple pseudo-speech-encoder configurations,
such as those using random duration models,
are surprisingly effective and can outperform more complex,
trainable versions.
However, there is still a significant gap
between this simple approach of text-utilization
and the performance of models utilizing synthetic speech from a TTS model.
The TTS model can be seen as a variant of the pseudo-speech-encoder model,
and we want to investigate the individual differences between both approaches in future work.

\ifcameraready
\section{Acknowledgements}

This work was partially supported by NeuroSys,
which as part of the initiative “Clusters4Future” is funded by the Federal Ministry of
Education and Research BMBF (funding IDs 03ZU2106DA and 03ZU2106DD),
and by the project RESCALE within the program
\textit{AI Lighthouse Projects for the Environment, Climate, Nature and Resources}
funded by
the Federal Ministry for the Environment, Nature Conservation,
Nuclear Safety and Consumer Protection (BMUV),
funding ID: 67KI32006A.
The authors gratefully acknowledge the computing time provided to them
at the NHR Center NHR4CES at RWTH Aachen University
(project number p0023565 and p0023999).
This is funded by the Federal Ministry of Education and Research,
and the state governments participating on the basis
of the resolutions of the GWK for
national high performance computing at universities
(\url{www.nhr-verein.de/unsere-partner}).
\fi

\section{Generative AI Use Disclosure}
We use LLMs to improve the formulations and grammar of the paper.

\bibliographystyle{IEEEtran}
\bibliography{mybib}

@misc{koch2025dlm,
      title={Reproducing and Dissecting Denoising Language Models for Speech Recognition}, 
      author={Dorian Koch and Albert Zeyer and Nick Rossenbach and Ralf Schlüter and Hermann Ney},
      year={2025},
      eprint={2512.13576},
      archivePrefix={arXiv},
      primaryClass={cs.NE},
      url={https://arxiv.org/abs/2512.13576}, 
}

@misc{gu2024lm,
  title={Denoising {LM}: Pushing the limits of error correction models for speech recognition},
  author={Gu, Zijin and Likhomanenko, Tatiana and Bai, He and McDermott, Erik and Collobert, Ronan and Jaitly, Navdeep},
  howpublished={arXiv:2405.15216},
  year={2024}
}

@inproceedings{panayotov2015librispeech,
  title={Librispeech: an {ASR} corpus based on public domain audio books},
  author={Panayotov, Vassil and Chen, Guoguo and Povey, Daniel and Khudanpur, Sanjeev},
  booktitle={2015 IEEE international conference on acoustics, speech and signal processing (ICASSP)},
  pages={5206--5210},
  year={2015},
  organization={IEEE}
}

@INPROCEEDINGS{zhou24cjst,
  author={Zhou, Wei and Jia, Junteng and Sari, Leda and Mahadeokar, Jay and Kalinli, Ozlem},
  booktitle={ICASSP 2025 - 2025 IEEE International Conference on Acoustics, Speech and Signal Processing (ICASSP)}, 
  title={{CJST}: {CTC} Compressor based Joint Speech and Text Training for Decoder-Only {ASR}}, 
  year={2025},
  volume={},
  number={},
  pages={1-5},
  doi={10.1109/ICASSP49660.2025.10888940}
  }

@inproceedings{gaido-etal-2021-ctc,
    title = "{CTC}-based Compression for Direct Speech Translation",
    author = "Gaido, Marco  and
      Cettolo, Mauro  and
      Negri, Matteo  and
      Turchi, Marco",
    editor = "Merlo, Paola  and
      Tiedemann, Jorg  and
      Tsarfaty, Reut",
    booktitle = "Proceedings of the 16th Conference of the European Chapter of the Association for Computational Linguistics: Main Volume",
    month = apr,
    year = "2021",
    address = "Online",
    publisher = "Association for Computational Linguistics",
    url = "https://aclanthology.org/2021.eacl-main.57/",
    doi = "10.18653/v1/2021.eacl-main.57",
    pages = "690--696"
 }

@inproceedings{gao22b_interspeech,
  title     = {Paraformer: Fast and Accurate Parallel Transformer for Non-autoregressive End-to-End Speech Recognition},
  author    = {Zhifu Gao and ShiLiang Zhang and Ian McLoughlin and Zhijie Yan},
  year      = {2022},
  booktitle = {{Interspeech 2022}},
  pages     = {2063--2067},
  doi       = {10.21437/Interspeech.2022-9996},
  issn      = {2958-1796},
}

@misc{an2024paraformerv2,
      title={Paraformer-v2: An improved non-autoregressive transformer for noise-robust speech recognition}, 
      author={Keyu An and Zerui Li and Zhifu Gao and Shiliang Zhang},
      year={2024},
      eprint={2409.17746},
      archivePrefix={arXiv},
      primaryClass={eess.AS},
      url={https://arxiv.org/abs/2409.17746}, 
}

@inproceedings{zou24_interspeech,
  title     = {{E-Paraformer}: A Faster and Better Parallel Transformer for Non-autoregressive End-to-End {Mandarin} Speech Recognition},
  author    = {Kun Zou and Fengyun Tan and Ziyang Zhuang and Chenfeng Miao and Tao Wei and Shaodan Zhai and Zijian Li and Wei Hu and Shaojun Wang and Jing Xiao},
  year      = {2024},
  booktitle = {{Interspeech 2024}},
  pages     = {267--271},
  doi       = {10.21437/Interspeech.2024-1891},
  issn      = {2958-1796},
}

@inproceedings{nozaki21_interspeech,
  title     = {Relaxing the Conditional Independence Assumption of {CTC}-Based {ASR} by Conditioning on Intermediate Predictions},
  author    = {Jumon Nozaki and Tatsuya Komatsu},
  year      = {2021},
  booktitle = {{Interspeech 2021}},
  pages     = {3735--3739},
  doi       = {10.21437/Interspeech.2021-911},
  issn      = {2958-1796},
}

@InProceedings { rossenbach:ICASSP2020,
author= {Rossenbach, Nick and Zeyer, Albert and Schlüter, Ralf and Ney, Hermann},
title= {Generating Synthetic Audio Data for Attention-Based Speech Recognition Systems},
booktitle= {IEEE International Conference on Acoustics, Speech, and Signal Processing},
year= 2020,
pages= {7069-7073},
address= {Barcelona, Spain},
month= may,
booktitlelink= {https://2020.ieeeicassp.org/}
}

@InProceedings{wang20222prefix-lm,
  title = 	 {What Language Model Architecture and Pretraining Objective Works Best for Zero-Shot Generalization?},
  author =       {Wang, Thomas and Roberts, Adam and Hesslow, Daniel and Scao, Teven Le and Chung, Hyung Won and Beltagy, Iz and Launay, Julien and Raffel, Colin},
  booktitle = 	 {Proceedings of the 39th International Conference on Machine Learning},
  pages = 	 {22964--22984},
  year = 	 {2022},
  editor = 	 {Chaudhuri, Kamalika and Jegelka, Stefanie and Song, Le and Szepesvari, Csaba and Niu, Gang and Sabato, Sivan},
  volume = 	 {162},
  series = 	 {Proceedings of Machine Learning Research},
  month = 	 {17--23 Jul},
  publisher =    {PMLR},
  url = 	 {https://proceedings.mlr.press/v162/wang22u.html}
}

@inproceedings{chen2022maestro,
  title     = {{MAESTRO}: Matched Speech Text Representations through Modality Matching},
  author    = {Zhehuai Chen and Yu Zhang and Andrew Rosenberg and Bhuvana Ramabhadran and Pedro J. Moreno and Ankur Bapna and Heiga Zen},
  year      = {2022},
  booktitle = {{Interspeech 2022}},
  pages     = {4093--4097},
  doi       = {10.21437/Interspeech.2022-10937},
  issn      = {2958-1796},
}

@inproceedings{sainath2023joist,
  title={{JOIST}: A joint speech and text streaming model for {ASR}},
  author={Sainath, Tara N and Prabhavalkar, Rohit and Bapna, Ankur and Zhang, Yu and Huo, Zhouyuan and Chen, Zhehuai and Li, Bo and Wang, Weiran and Strohman, Trevor},
  booktitle={2022 IEEE spoken language technology workshop (SLT)},
  pages={52--59},
  year={2023},
  organization={IEEE}
}

@inproceedings{zhang2022speechut,
  title={{SpeechUT}: Bridging speech and text with hidden-unit for encoder-decoder based speech-text pre-training},
  author={Zhang, Ziqiang and Zhou, Long and Ao, Junyi and Liu, Shujie and Dai, Lirong and Li, Jinyu and Wei, Furu},
  booktitle={Proceedings of the 2022 Conference on Empirical Methods in Natural Language Processing},
  pages={1663--1676},
  year={2022}
}

@inproceedings{wang2023understandingsharedrepr,
  title={Understanding shared speech-text representations},
  author={Wang, Gary and Kastner, Kyle and Bapna, Ankur and Chen, Zhehuai and Rosenberg, Andrew and Ramabhadran, Bhuvana and Zhang, Yu},
  booktitle={ICASSP 2023-2023 IEEE International Conference on Acoustics, Speech and Signal Processing (ICASSP)},
  pages={1--5},
  year={2023},
  organization={IEEE}
}

@INPROCEEDINGS{Sunder2025nonautoregressive,
  author={Sunder, Vishal and Kingsbury, Brian and Saon, George and Thomas, Samuel and Shechtman, Slava and Aronowitz, Hagai and Fosler-Lussier, Eric and Lastras, Luis},
  booktitle={ICASSP 2025 - 2025 IEEE International Conference on Acoustics, Speech and Signal Processing (ICASSP)}, 
  title={A Non-autoregressive Model for Joint {STT} and {TTS}}, 
  year={2025},
  volume={},
  number={},
  pages={1-5},
  doi={10.1109/ICASSP49660.2025.10887605}}

@INPROCEEDINGS{thomas2022textogram,
  author={Thomas, Samuel and Kuo, Hong-Kwang J. and Kingsbury, Brian and Saon, George},
  booktitle={ICASSP 2022 - 2022 IEEE International Conference on Acoustics, Speech and Signal Processing (ICASSP)}, 
  title={Towards Reducing the Need for Speech Training Data to Build Spoken Language Understanding Systems}, 
  year={2022},
  volume={},
  number={},
  pages={7932-7936},
  doi={10.1109/ICASSP43922.2022.9747555}}

@inproceedings{ao-etal-2022-speecht5,
    title = "{SpeechT5}: Unified-Modal Encoder-Decoder Pre-Training for Spoken Language Processing",
    author = "Ao, Junyi  and
      Wang, Rui  and
      Zhou, Long  and
      Wang, Chengyi  and
      Ren, Shuo  and
      Wu, Yu  and
      Liu, Shujie  and
      Ko, Tom  and
      Li, Qing  and
      Zhang, Yu  and
      Wei, Zhihua  and
      Qian, Yao  and
      Li, Jinyu  and
      Wei, Furu",
    editor = "Muresan, Smaranda  and
      Nakov, Preslav  and
      Villavicencio, Aline",
    booktitle = "Proceedings of the 60th Annual Meeting of the Association for Computational Linguistics (Volume 1: Long Papers)",
    month = may,
    year = "2022",
    address = "Dublin, Ireland",
    publisher = "Association for Computational Linguistics",
    url = "https://aclanthology.org/2022.acl-long.393/",
    doi = "10.18653/v1/2022.acl-long.393",
    pages = "5723--5738"

}

@misc{bapna2021slamunifiedencoderspeech,
      title={{SLAM}: A Unified Encoder for Speech and Language Modeling via Speech-Text Joint Pre-Training}, 
      author={Ankur Bapna and Yu-an Chung and Nan Wu and Anmol Gulati and Ye Jia and Jonathan H. Clark and Melvin Johnson and Jason Riesa and Alexis Conneau and Yu Zhang},
      year={2021},
      eprint={2110.10329},
      archivePrefix={arXiv},
      primaryClass={cs.CL},
      url={https://arxiv.org/abs/2110.10329}, 
}

@misc{bapna2022mslammassivelymultilingualjoint,
      title={{mSLAM}: Massively multilingual joint pre-training for speech and text}, 
      author={Ankur Bapna and Colin Cherry and Yu Zhang and Ye Jia and Melvin Johnson and Yong Cheng and Simran Khanuja and Jason Riesa and Alexis Conneau},
      year={2022},
      eprint={2202.01374},
      archivePrefix={arXiv},
      primaryClass={cs.CL},
      url={https://arxiv.org/abs/2202.01374}, 
}

@inproceedings{tang-etal-2022-unified,
    title = "Unified Speech-Text Pre-training for Speech Translation and Recognition",
    author = "Tang, Yun  and
      Gong, Hongyu  and
      Dong, Ning  and
      Wang, Changhan  and
      Hsu, Wei-Ning  and
      Gu, Jiatao  and
      Baevski, Alexei  and
      Li, Xian  and
      Mohamed, Abdelrahman  and
      Auli, Michael  and
      Pino, Juan",
    editor = "Muresan, Smaranda  and
      Nakov, Preslav  and
      Villavicencio, Aline",
    booktitle = "Proceedings of the 60th Annual Meeting of the Association for Computational Linguistics (Volume 1: Long Papers)",
    month = may,
    year = "2022",
    address = "Dublin, Ireland",
    publisher = "Association for Computational Linguistics",
    url = "https://aclanthology.org/2022.acl-long.105/",
    doi = "10.18653/v1/2022.acl-long.105",
    pages = "1488--1499"
}

@inproceedings{chung-etal-2021-splat,
    title = "{SPLAT}: Speech-Language Joint Pre-Training for Spoken Language Understanding",
    author = "Chung, Yu-An  and
      Zhu, Chenguang  and
      Zeng, Michael",
    editor = "Toutanova, Kristina  and
      Rumshisky, Anna  and
      Zettlemoyer, Luke  and
      Hakkani-Tur, Dilek  and
      Beltagy, Iz  and
      Bethard, Steven  and
      Cotterell, Ryan  and
      Chakraborty, Tanmoy  and
      Zhou, Yichao",
    booktitle = "Proceedings of the 2021 Conference of the North American Chapter of the Association for Computational Linguistics: Human Language Technologies",
    month = jun,
    year = "2021",
    address = "Online",
    publisher = "Association for Computational Linguistics",
    url = "https://aclanthology.org/2021.naacl-main.152/",
    doi = "10.18653/v1/2021.naacl-main.152",
    pages = "1897--1907"
}

@inproceedings{gaur24astra,
  title     = {{ASTRA}: Aligning Speech and Text Representations for {ASR} without Sampling},
  author    = {Neeraj Gaur and Rohan Agrawal and Gary Wang and Parisa Haghani and Andrew Rosenberg and Bhuvana Ramabhadran},
  year      = {2024},
  booktitle = {{Interspeech 2024}},
  pages     = {3904--3908},
  doi       = {10.21437/Interspeech.2024-1924},
  issn      = {2958-1796},
}

@INPROCEEDINGS{dalterio2023tts,
  author={D’Alterio, Pasquale and Hensel, Christian and Hasan, Bashar Awwad Shiekh},
  booktitle={2023 IEEE Automatic Speech Recognition and Understanding Workshop (ASRU)}, 
  title={Can Unpaired Textual Data Replace Synthetic Speech in {ASR} Model Adaptation?}, 
  year={2023},
  volume={},
  number={},
  pages={1-8},
  doi={10.1109/ASRU57964.2023.10389722}}

\end{document}